\documentclass{bmvc2k}

\title{Weakly-supervised Localization of Manipulated Image Regions Using Multi-resolution Learned Features}

\addauthor{Ziyong Wang}{zwang172@sheffield.ac.uk}{1}
\addauthor{Charith Abhayaratne}{c.abhayaratne@sheffield.ac.uk}{1}

\addinstitution{
 Department of Electronic and Electrical Engineering,\\
 The University of Sheffield,\\
 S1 3JD Sheffield, United Kingdom
}

\runninghead{Wang, Abhayaratne}{Weakly-supervised Manipulation localization}

\newcommand{\mywidthS}{0.14\textwidth}
\newcommand{\mywidthSS}{0.13\textwidth}

\usepackage{amsmath}
\usepackage{multirow}
\usepackage{graphicx}
\usepackage{amssymb} 
\usepackage{amsmath}
\usepackage{algorithm}
\usepackage{algpseudocode}
\usepackage{wrapfig}
\begin{document}

\maketitle

\begin{abstract}
The explosive growth of digital images and the widespread availability of image editing tools have made image manipulation detection an increasingly critical challenge. Current deep learning-based manipulation detection methods excel in achieving high image-level classification accuracy, they often fall short in terms of interpretability and localization of manipulated regions. Additionally, the absence of pixel-wise annotations in real-world scenarios limits the existing fully-supervised manipulation localization techniques. To address these challenges, we propose a novel weakly-supervised approach that integrates activation maps generated by image-level manipulation detection networks with segmentation maps from pre-trained models. Specifically, we build on our previous image-level work named WCBnet to produce multi-view feature maps which are subsequently fused for coarse localization. These coarse maps are then refined using detailed segmented regional information provided by pre-trained segmentation models (such as DeepLab, SegmentAnything and PSPnet), with Bayesian inference employed to enhance the manipulation localization. Experimental results demonstrate the effectiveness of our approach, highlighting the feasibility to localize image manipulations without relying on pixel-level labels.  
\end{abstract}

\section{Introduction}
\label{sec:intro}
With the rapid growth and increasing accessibility of image manipulation tools, manipulated images are being produced at an alarming rate, posing a significant threat to cyber security~\cite{thakur2020recent,survey1}. Effective detection of manipulated images is vital for maintaining the integrity of visual information across domains such as journalism, legal evidence, and social media~\cite{zanardelli2023image,tyagi2023detailed}. In recent years, deep learning has been commonly used for image manipulation detection which enables remarkably high performance at either image-level classification~\cite{camacho2022convolutional,hamid2023improvised} or pixel-level localization~\cite{Guo_2023_CVPR,9298821}. However in real world, the manipulated images typically do not have manually annotated pixel-level labels. It prevents the image-level methods from being further leveraged to precisely localize manipulated regions and provide sufficient interpretability~\cite{walia2022using,abir2023detecting}. The absence of annotated data also restricts the effectiveness of current fully-supervised manipulation localization methods~\cite{zeng2024semi}. Thus, the challenge of accurately localizing manipulated regions within images without pixel-level labels remains challenging.

Several weakly-supervised image manipulation localization methods have been proposed in recent years. The activation maps of image-wise manipulation model, extracted by grad-CAM, are directly applied as pixel-wise predictions and compared with semi- or fully-supervised models~\cite{tantaru2024weakly}. Object edges extracted by clustering super-pixels are applied to enhance the grad-CAM result in order to localize the manipulated regions~\cite{zhou2024exploring}, while another method applies a three-source stream (RGB, SRM and Bayar) to generate pseudo pixel-wise labels leveraging the image-wise model for manipulation localization~\cite{zhai2023towards}. However, enhancement through image segmentation is more closely aligned with the nature of image manipulation itself, as it involves the insertion of meaningful objects or regions to create misinformation~\cite{zhang2020survey}. Segmentation-based approaches~\cite{haciefendiouglu2024automatic,Hasany_2023_CVPR,kweon2023weakly} can provide pixel-level details but typically require extensive labelled data, which is often unavailable. Consequently, the segmentation models require collaboration with image manipulation detection methods for distinguishing the manipulated class.

This paper addresses the challenge of localizing image manipulations without pixel-level annotations by proposing a novel method that combines the multi-view activation map of the classification network with the fine-grained region captures of pre-trained segmentation models. Specifically, our image-wise manipulation network is built on a structure named Cross-block Attention Module (CBAM) from our previous work, an image-wise manipulation method WCBnet~\cite{wang2023wcbnet}, which weights and fuses the convolutional block output feature at single fixed receptive field. The multi-view activation map is obtained by computing grad-CAMs~\cite{selvaraju2020grad} on differently-fused features from varying configurations of CBAMs, the geometric mean of which involves multi-resolution activation maps across comprehensive receptive fields. This multi-view activation map is considered as coarse localization, which is further leveraged by several pre-trained segmentation models, such as DeepLab~\cite{deeplab2_2021}, SAM~\cite{kirillov2023segany}, and PSPnet~\cite{gupta2023image} that segment the images into potentially-manipulated regions. Eventually, the activation maps are integrated with these segmentation maps via Bayesian inference, thereby generating more accurate manipulation localization results. The primary contribution of this work is the demonstration of the feasibility for accurate image manipulation localization without requiring pixel-level annotations. This is accomplished by the innovative combination of activation maps from image classification networks and region masks from pre-trained segmentation networks.

\section{Methodology}
\label{sec:method}

\begin{figure*}
\begin{center}
\includegraphics[width = 0.9\textwidth]{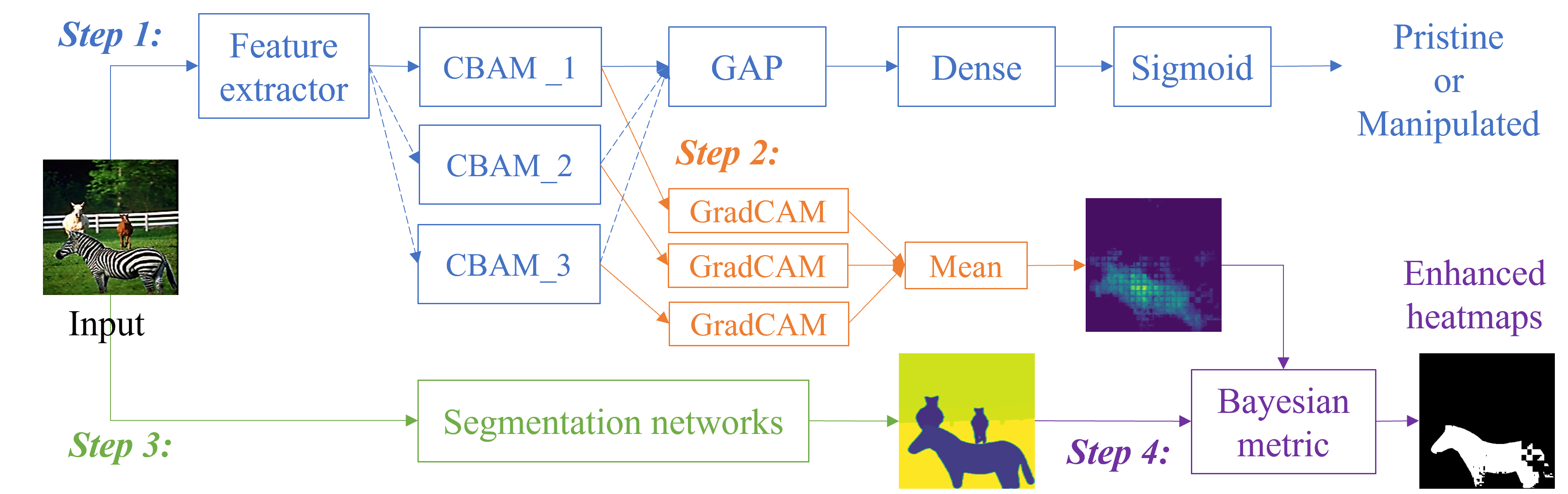}
\end{center}
   \caption{The work-flow of proposed weakly-supervised manipulation localization model; Each step is labeled and presented in different colors. }
\label{fig:WSWCBnet}
\end{figure*}

In this section, we describe the methodology employed to detect image manipulation and enhance the localization of manipulated regions without using pixel-wise ground-truth labels. As shown in the Figure~\ref{fig:WSWCBnet}, our approach consists of four main steps: image-wise manipulation classification, feature map generation, segmentation map extraction using a pre-trained network, and combining these outputs to produce enhanced heatmaps.

\subsection{Image-wise Image Manipulation Classification}
\label{subsec:classification}

The first step in our method involves classifying whether an image has been manipulated or pristine. In this study, a CNN-based feature extractor is employed to extract the hierarchical features from input images which contain global and local information of manipulation traces. Since both types of information are equally important in image manipulation detection, we apply Cross-block Attention Module (CBAM) from an image manipulation detection network (called WCBnet from our previous work~\cite{wang2023wcbnet}) to assign adaptive weights for these high-level and low-level features and fuse them. Specifically, the input three-channel (RGB) image of shape \( x \in \mathbb{R}^{384 \times 384 \times 3} \) is first processed by a CNN-based feature extractor (ResNet50) to produce multi-scale feature maps \( \{F_i\}_{i=1}^5 \), where each \( F_i \) represents features extracted by 5 convolutional blocks at different scales. \( F_i \in \mathbb{R}^{H_i \times W_i \times K_i} \) that represents the feature map at scale \( i \) is defined as:
\begin{equation}
	F_i = \text{CNN}(x).
\label{eq:CNN}
\end{equation}

These feature maps are then fed into the CBAM structure, which rescales the features to a consistent dimensionality \( \mathbb{R}^{H \times W \times D} \) and applies self-attention mechanisms to model the relationships between different convolutional blocks. This process assigns trainable weights according to the inner relationships between features, enhancing the model's ability to capture and utilize complex feature dependencies. These feature maps are weighted and concatenated as: 
\begin{equation}
F_c = \text{CBAM}(F_1, F_2, \ldots, F_N).
\label{eq:CBWM}
\end{equation}

The processed features \( F_c \) are then passed through a global average pooling (GAP) layer to obtain a fixed-size feature vector \( v \in \mathbb{R}^D \). This vector is subsequently input to a dense layer, producing logits \( z \in \mathbb{R}^C \), where \( C \) is the number of classes. Finally, a Sigmoid activation function \( \sigma(z) \) is applied to the logits to yield binary classification probabilities \( p \in [0, 1] \) belonging to the manipulated class.

\subsection{Fusion of Multi-Scale Activation Maps}
\label{subsec:gradcam}

Upon classifying the image as manipulated, we apply Gradient-weighted Class Activation Mapping (Grad-CAM~\cite{selvaraju2020grad}) to identify and visualize the model's region of interest that responds actively when classifying manipulated images. The activation map \( A\) is computed by evaluating the gradients between the weighted feature maps \( F_c \) and the network's output. Furthermore, we incorporate multi-view activation maps \( A_i \), derived from distinct feature sets \( F_c \), which are produced by varying configurations of the Cross-block Attention Module (CBAM). This module effectively weights and fuses features \( F_i \) across multiple scales, enhancing the feature representation of manipulation-related traces. In this paper, we compute Grad-CAM activation maps \( A_i \) for scales \( i = 2, 3, 4 \) which correspond to the receptive field of each feature map \( F_2 \), \( F_3 \) and \( F_4 \) that contains proper levels of information. These multi-scale activation maps are then aggregated by geometric mean as:
\begin{equation}
A = \left( \prod_{i \in \{2, 3, 4\}} A_i \right)^{\frac{1}{n}},
\end{equation}
where \( n = 3 \) is the number of scales used for the geometric mean computation. This approach effectively combines the detailed activation maps with global context, enhancing the detection and visualization of manipulated regions.

\subsection{Generating Segmentation Maps Using Pre-trained Models}
\label{subsec:segmentation}
To further enhance the precision of the activation maps \( A \) generated by the image-wise manipulation classification model, we integrate several pre-trained segmentation models to divide the image into distinct regions that potentially contain manipulated areas. The segmentation models employed, namely DeepLab2~\cite{deeplab2_2021}, the Segmentation Anything Model (SAM)~\cite{kirillov2023segany} and PSPnet~\cite{gupta2023image}, have been pre-trained on large-scale datasets to detect and segment normal objects. These models offer pixel-level accuracy in identifying potential manipulation regions with higher precision. The resulting segmentation maps \( M \), which consist of various masks \(M_i\) corresponding to each detected object \(i\), are defined as follows:
\begin{equation}
M = \bigcup_{i=1}^{n} M_i.
\end{equation}
However, the presence of numerous detectable objects within an image makes it difficult for segmentation networks to distinguish manipulated regions from the array of detected objects.
\subsection{Combining Activation Maps and Segmentation Maps}
\label{subsec:bayesian}

In the final step, we combine the activation maps \( A\) that shows the coarse regions of interest for image-wise manipulation classification, and the segmentation maps \( M\) illustrate delicate regions of massive objects without semantic information. To enhance the multi-view activation map \( A \) using the segmentation map \( M \), we first compute the similarity between \( A\) and each object region mask \( M_i \) in \( m \) by the function \( S(\cdot) \).  For this manipulation task, we utilize the distance transform \( D \) which measures the proximity of each pixel in \( A \) to the nearest boundary of \( M_i \) to compute the similarity score. The weighted sum of each pixel value multiplied by the distance to the region edge is defined as:
\begin{equation}
S(M_i, A) = \frac{1}{\sum M_i} \sum (D(M_i) \cdot A).
\label{eq:similarity}
\end{equation}
In the above equation~\ref{eq:similarity}, we normalize the result of each region \( M_i \) by its size $\sum M_i$ to ensure that smaller regions have a fair impact on the similarity score. This metric effectively captures how closely the predicted manipulated regions align with the segmented object boundaries, providing a statistical measure of spatial accuracy. We then identify the mask \( M_{i^*} \) that maximizes this similarity which is subsequently used to enhance the activation map \( A \), producing the refined manipulation heatmap \( H\), defined as:
\begin{equation}
A^* = E(A, M_{i^*}), \quad \text{where } i^* = \arg\max_{i} S(M_i, A).
\end{equation}
In our method, the function \( E(\cdot) \) is Bayesian inference that enhances the activation map \( A\) incorporating additional information from the most similar binary segmentation mask \( M_{i^*} \). Here, \( P(A) \) represents the manipulation probability of the activation map \( A \), \( P(M_{i^*}) \) is the prior probability of the mask \( M_{i^*} \), and \( P(A \mid M_{i^*}) \) denotes the conditional probability of \( A \) given the mask \( M_{i^*} \). The enhanced activation heatmap \( A^* \),  is computed as follows:
\begin{equation}
P(A \mid M_{i^*}) = \frac{P(M_{i^*} \mid  A) \cdot P(A)}{P(M_{i^*})}.
\end{equation}
This Bayesian inference refines the initial coarse activation map by incorporating spatial information of segmentation maps.

The presented method combines image manipulation classification, weakly-supervised localization, and segmentation techniques to achieve the detection and localization of manipulated regions in the absence of pixel-wise labels. The integration of multi-view activation maps and pre-trained segmentation networks via Bayesian inference, combining the semantic information from the manipulation classification network and the fine-grained regional information from segmentation networks into enhanced manipulation localization results.

%
%
%


\section{Experimental Results}
\label{sec:experiment}
In this section, we present a series of experiments designed to evaluate the effectiveness of our proposed multi-view activation map approach, assess different segmentation networks, and examine the performance of enhanced heatmaps generated by combining activation maps with segmentation masks. Before that, the experimental setup of evaluating our proposed model on the task of generating enhanced heatmap of image manipulation at the absence of pixel-wise labels is illustrated.
\subsection{Dataset}
We conduct our experiments mainly on CASIA2.0 image manipulation dataset~\cite{6625374}, and select approximately 1800 splicing, 1800 copy-move and 1800 authentic images to train the image-wise feature extractor (WCBnet) and generate multi-view activation maps. Each image is resized to 384x384 pixels for consistency. All images are pre-processed using signed-value error levels~\cite{gunawan2017development} following the experimental setup of WCBnet, to extract JPEG compression-based artifacts of manipulation traces.
\subsection{Experimental Setup}
The experiments were conducted on a server with NVIDIA GeForce RTX 3090 Ti, 12th Gen Intel (R) Core (TM) i9-12900K 3.20 GHz processor and 32.0 GB RAM, and the model is constructed based on Python 3.7 and TensorFlow 2.7. For the image-wise classification step, our model is based on the ResNet-50 architecture, with additional Cross Convolutional-blocks Weighting module for feature weighting and fusion on the image manipulation tasks. The image-wise model was trained using the SGD optimizer with an initial learning rate of 0.001, a batch size of 12, and for 200 epochs. For the segmentation models, several state-of-art image segmentation models, namely DeepLab2, Segmentation Anything Model (SAM) and PSPnet which have been trained on object detection datasets are used to generate pixel-wise masks for each region. To claim, we only use the segmentation masks from DeepLab although it could be a semantic segmentation model. The image-wise performance is evaluated using accuracy and F1-score to account for the class imbalance in the dataset, while the pixel-wise performance is evaluated using Area Under the Curve (AUC) and F1-score with fixed threshold. We conduct experiments to visually and statistically prove the effectiveness of multi-view activation maps and the combination with segmentation masks. 
\subsection{Multi-view Activation Map Fusion}
\begin{figure}[t]
\centering
\subfigure{\includegraphics[width = \mywidthS , height = \mywidthS]{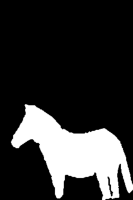}}
\subfigure{\includegraphics[width = \mywidthS, height= \mywidthS]{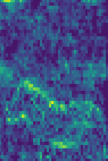}}
\subfigure{\includegraphics[width = \mywidthS , height = \mywidthS]{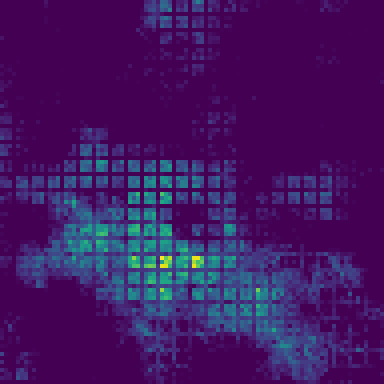}}
\subfigure{\includegraphics[width = \mywidthS , height = \mywidthS]{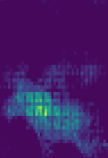}}
\subfigure{\includegraphics[width = \mywidthS , height = \mywidthS]{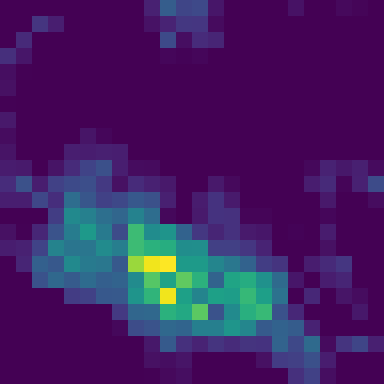}}
\subfigure{\includegraphics[width = \mywidthS , height = \mywidthS]{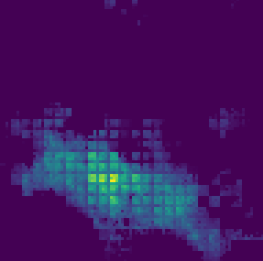}}

\subfigure{\includegraphics[width = \mywidthS , height = \mywidthS]{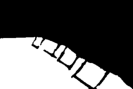}}
\subfigure{\includegraphics[width = \mywidthS , height = \mywidthS]{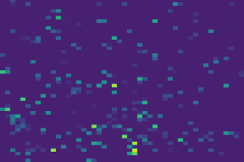}}
\subfigure{\includegraphics[width = \mywidthS , height = \mywidthS]{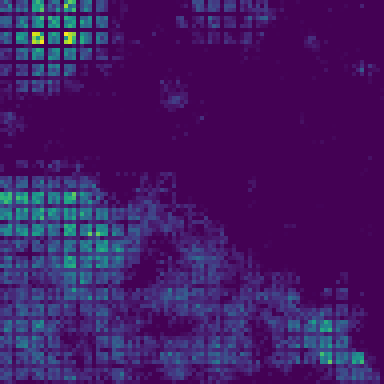}}
\subfigure{\includegraphics[width = \mywidthS , height = \mywidthS]{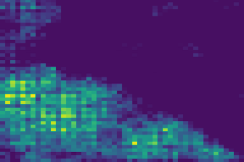}}
\subfigure{\includegraphics[width = \mywidthS , height = \mywidthS]{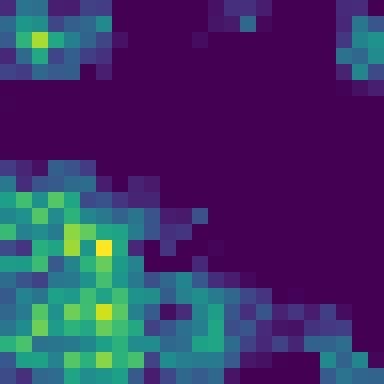}}
\subfigure{\includegraphics[width = \mywidthS , height = \mywidthS]{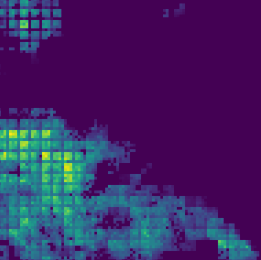}}

\subfigure{\includegraphics[width = \mywidthS , height = \mywidthS]{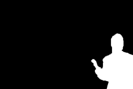}}
\subfigure{\includegraphics[width = \mywidthS , height = \mywidthS]{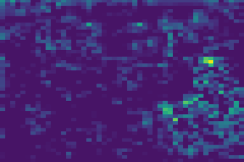}}
\subfigure{\includegraphics[width = \mywidthS , height = \mywidthS]{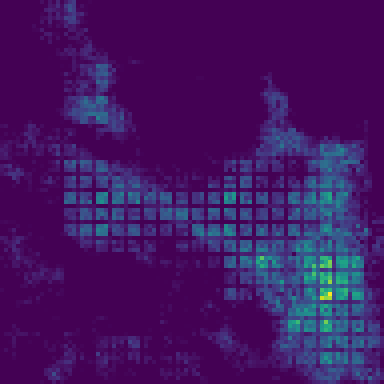}}
\subfigure{\includegraphics[width = \mywidthS , height = \mywidthS]{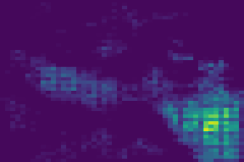}}
\subfigure{\includegraphics[width = \mywidthS , height = \mywidthS]{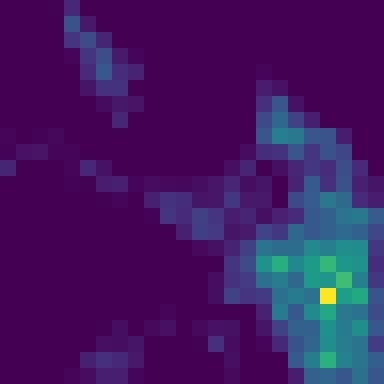}}
\subfigure{\includegraphics[width = \mywidthS , height = \mywidthS]{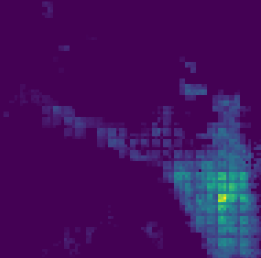}}

\subfigure{\includegraphics[width = \mywidthS , height = \mywidthS]{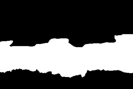}}
\subfigure{\includegraphics[width = \mywidthS , height = \mywidthS]{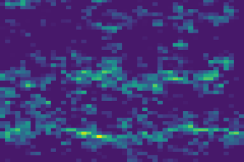}}
\subfigure{\includegraphics[width = \mywidthS , height = \mywidthS]{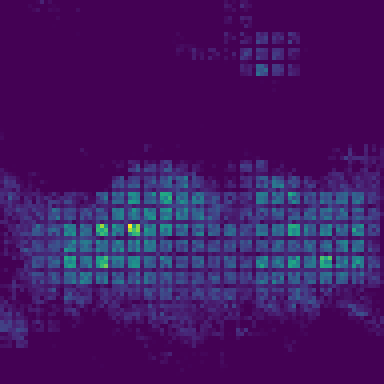}}
\subfigure{\includegraphics[width = \mywidthS , height = \mywidthS]{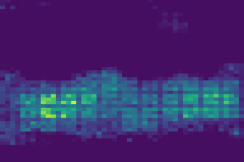}}
\subfigure{\includegraphics[width = \mywidthS , height = \mywidthS]{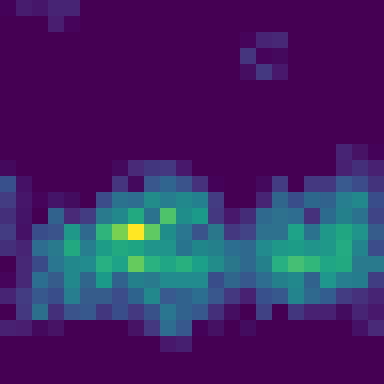}}
\subfigure{\includegraphics[width = \mywidthS , height = \mywidthS]{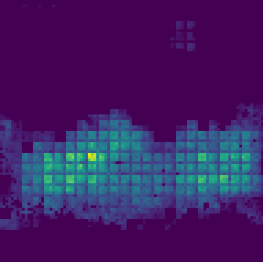}}

\setcounter{subfigure}{0}
\subfigure[Ref label]{\includegraphics[width = \mywidthS , height = \mywidthS]{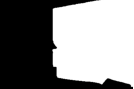}}
\subfigure[ResNet50]{\includegraphics[width = \mywidthS , height = \mywidthS]{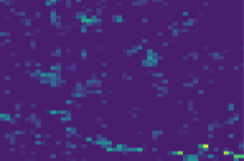}}
\subfigure[$\text{WCBnet}_2$]{\includegraphics[width = \mywidthS , height = \mywidthS]{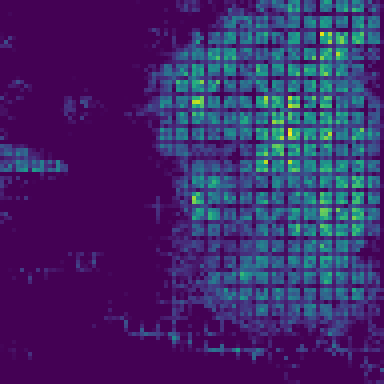}}
\subfigure[$\text{WCBnet}_3$]{\includegraphics[width = \mywidthS , height = \mywidthS]{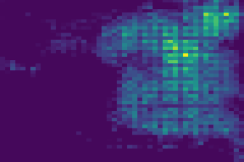}}
\subfigure[$\text{WCBnet}_4$]{\includegraphics[width = \mywidthS , height = \mywidthS]{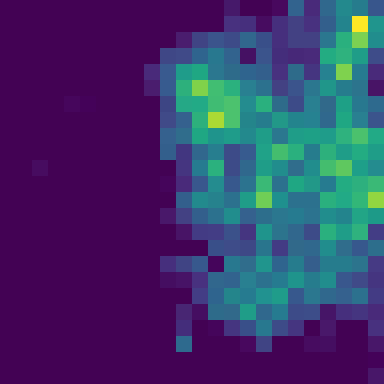}}
\subfigure[$\text{WCBnet}_m$]{\includegraphics[width = \mywidthS , height = \mywidthS]{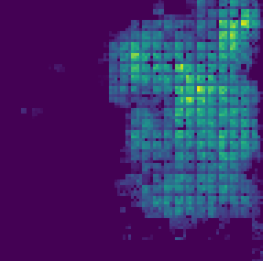}}

\caption{The features maps of the backbone ResNet50 and WCBnet; The $\text{WCBnet}_i$ means the CBAM with shape of block i , while $\text{WCBnet}_m$ is the their geometric mean; Ref label is the pixel-wise labels just for reference.}
\label{fig:CAMs}
\end{figure}
For the image-wise manipulation detection model (called WCBnet) applied in the paper, it employed single single-view CBAM structure to weight and fuse convolutional blocks within a fixed receptive field, achieving outstanding image-level manipulation classification accuracy across multiple datasets. Building upon this, our model investigates the impact of multiple CBAM structures regarding massive receptive fields and the resulting activation maps on manipulation localization. Specifically, we train several variants of WCBnet, namely, \(\text{WCBnet}_2\), \(\text{WCBnet}_3\), and \(\text{WCBnet}_4\), each incorporating different CBAM configurations. These models achieve commendable F1-scores of 0.912, 0.933, and 0.935, respectively, for image-level manipulation classification tasks on CASIA2.0 dataset. Additionally, by computing class activation maps between the weighted feature layers of these \(\text{WCBnet}_i\) models and their output layers, we extract feature maps that highlight the most active pixels during classification.

As illustrated in Figure~\ref{fig:CAMs}, the backbone ResNet50 exhibits limited focus on the manipulated regions, despite correctly classifying the image type. In contrast, the feature maps from WCBnet clearly delineate the boundaries between manipulated and background regions. However, the activation maps produced by shallow CBAM exhibit excessive highlighting of background regions, while those from deeper CBAM lack detailed information nearby region edges. The geometric mean of these activation maps provides a more accurate representation of the manipulation regions.

\subsection{Segmentation Network Comparison on Manipulated Images}
\begin{figure}[t]
\centering
\subfigure{\includegraphics[width=\mywidthS, height=\mywidthS]{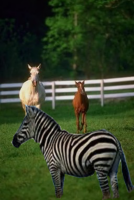}}
\subfigure{\includegraphics[width=\mywidthS, height=\mywidthS]{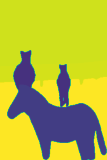}}
\subfigure{\includegraphics[width=\mywidthS, height=\mywidthS]{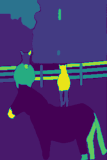}}
\subfigure{\includegraphics[width=\mywidthS, height=\mywidthS]{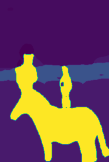}}

\subfigure{\includegraphics[width=\mywidthS, height=\mywidthS]{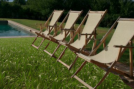}}
\subfigure{\includegraphics[width=\mywidthS, height=\mywidthS]{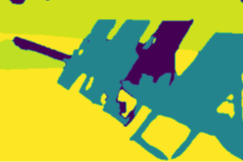}}
\subfigure{\includegraphics[width=\mywidthS, height=\mywidthS]{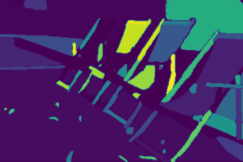}}
\subfigure{\includegraphics[width=\mywidthS, height=\mywidthS]{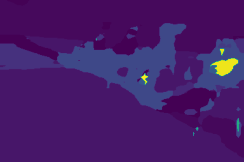}}

\subfigure{\includegraphics[width=\mywidthS, height=\mywidthS]{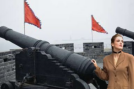}}
\subfigure{\includegraphics[width=\mywidthS, height=\mywidthS]{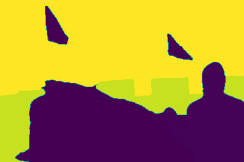}}
\subfigure{\includegraphics[width=\mywidthS, height=\mywidthS]{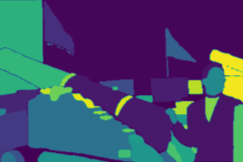}}
\subfigure{\includegraphics[width=\mywidthS, height=\mywidthS]{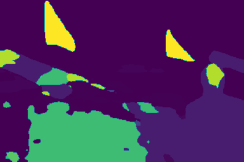}}

\subfigure{\includegraphics[width=\mywidthS, height=\mywidthS]{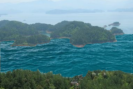}}
\subfigure{\includegraphics[width=\mywidthS, height=\mywidthS]{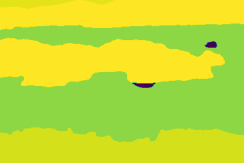}}
\subfigure{\includegraphics[width=\mywidthS, height=\mywidthS]{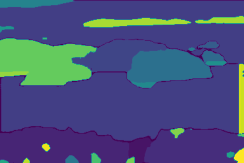}}
\subfigure{\includegraphics[width=\mywidthS, height=\mywidthS]{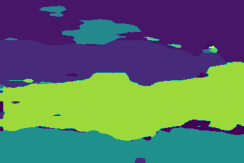}}

\setcounter{subfigure}{0}
\subfigure[Input]{\includegraphics[width=\mywidthS, height=\mywidthS]{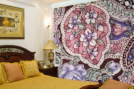}}
\subfigure[DeepLab]{\includegraphics[width=\mywidthS, height=\mywidthS]{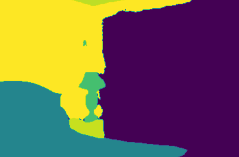}}
\subfigure[SAM]{\includegraphics[width=\mywidthS, height=\mywidthS]{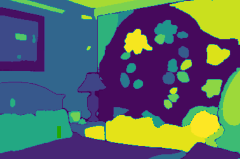}}
\subfigure[PSPnet]{\includegraphics[width=\mywidthS, height=\mywidthS]{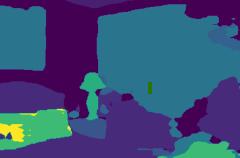}}

\caption{The manipulated images and their corresponding image segmentation maps, generated by three state-of-art pre-trained methods.}
\label{fig:Segs}
\end{figure}
As previously discussed, we utilized three state-of-the-art image segmentation models with pre-trained weights (DeepLab, SAM, and PSPNet) to detect and localize potentially manipulated regions within the images. Several samples of manipulated images along with their corresponding segmentation maps produced by these models are visualized for comparison.

As shown in the Figure~\ref{fig:Segs}, DeepLab is effective for segmenting large areas and images with a limited number of categories. In contrast, PSPNet and SAM excel at segmenting a broader range of smaller regions within the image. However, DeepLab may struggle to distinguish between manipulated and adjacent regions, while PSPNet and SAM may detect overly small areas, potentially missing parts of the manipulated object. These segmentation maps require to be combined with class activation maps to distinguish the manipulated regions.
\subsection{Combining Multi-view Feature Maps and Segmentation Masks}
\begin{figure}[t]
\centering
\subfigure{\includegraphics[width=\mywidthSS, height=\mywidthSS]{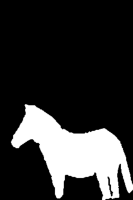}}
\subfigure{\includegraphics[width=\mywidthSS, height=\mywidthSS]{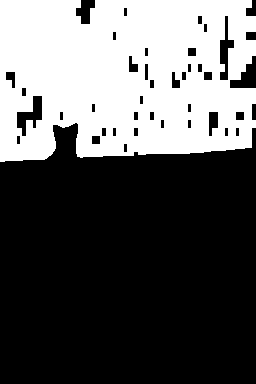}}
\subfigure{\includegraphics[width=\mywidthSS, height=\mywidthSS]{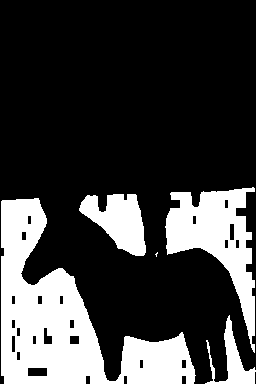}}
\subfigure{\includegraphics[width=\mywidthSS, height=\mywidthSS]{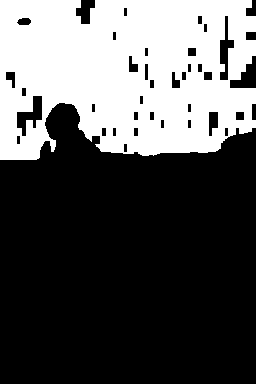}}
\subfigure{\includegraphics[width=\mywidthSS, height=\mywidthSS]{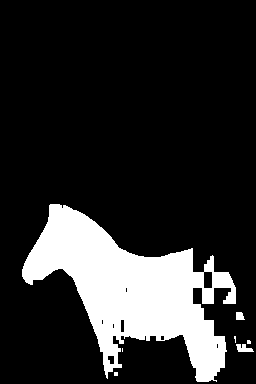}}
\subfigure{\includegraphics[width=\mywidthSS, height=\mywidthSS]{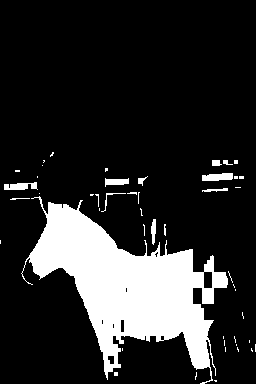}}
\subfigure{\includegraphics[width=\mywidthSS, height=\mywidthSS]{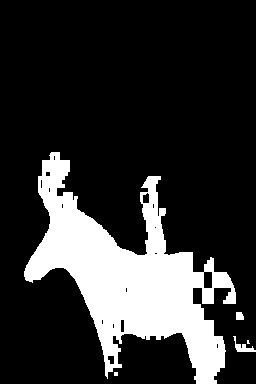}}

\subfigure{\includegraphics[width=\mywidthSS, height=\mywidthSS]{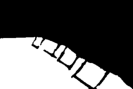}}
\subfigure{\includegraphics[width=\mywidthSS, height=\mywidthSS]{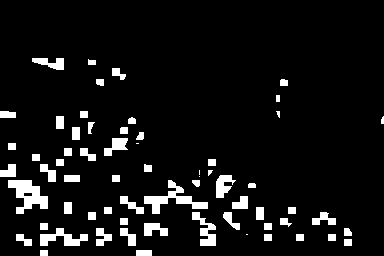}}
\subfigure{\includegraphics[width=\mywidthSS, height=\mywidthSS]{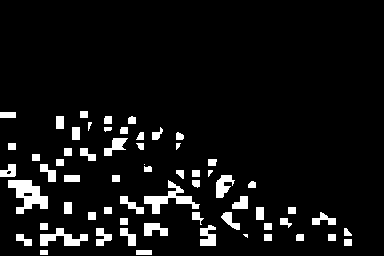}}
\subfigure{\includegraphics[width=\mywidthSS, height=\mywidthSS]{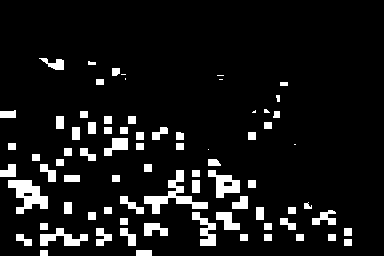}}
\subfigure{\includegraphics[width=\mywidthSS, height=\mywidthSS]{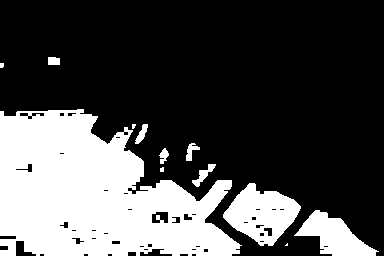}}
\subfigure{\includegraphics[width=\mywidthSS, height=\mywidthSS]{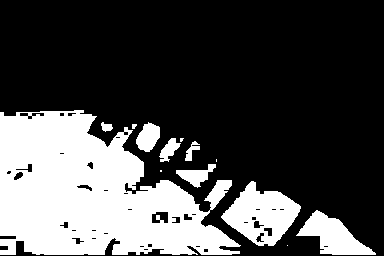}}
\subfigure{\includegraphics[width=\mywidthSS, height=\mywidthSS]{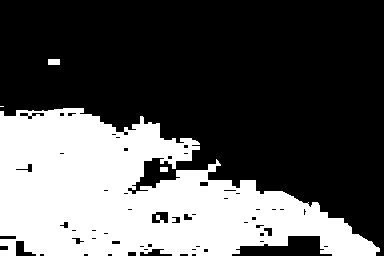}}

\subfigure{\includegraphics[width=\mywidthSS, height=\mywidthSS]{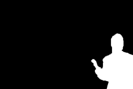}}
\subfigure{\includegraphics[width=\mywidthSS, height=\mywidthSS]{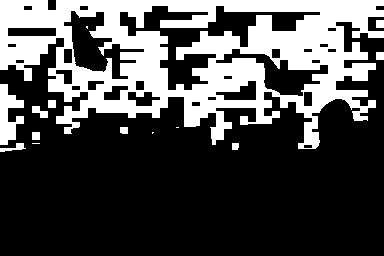}}
\subfigure{\includegraphics[width=\mywidthSS, height=\mywidthSS]{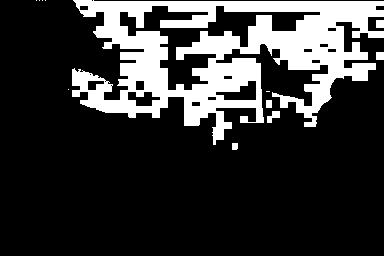}}
\subfigure{\includegraphics[width=\mywidthSS, height=\mywidthSS]{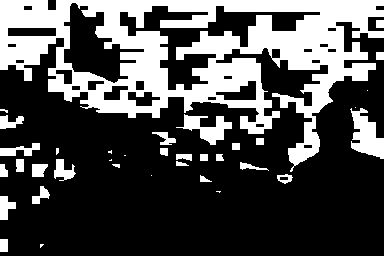}}
\subfigure{\includegraphics[width=\mywidthSS, height=\mywidthSS]{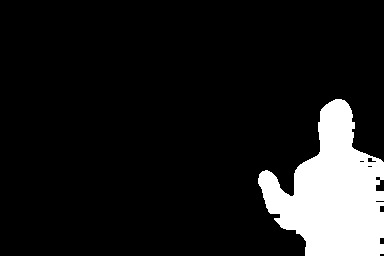}}
\subfigure{\includegraphics[width=\mywidthSS, height=\mywidthSS]{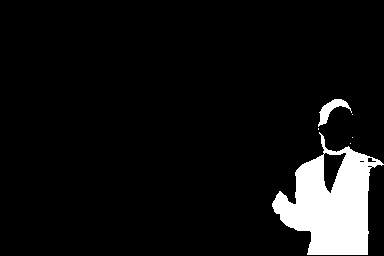}}
\subfigure{\includegraphics[width=\mywidthSS, height=\mywidthSS]{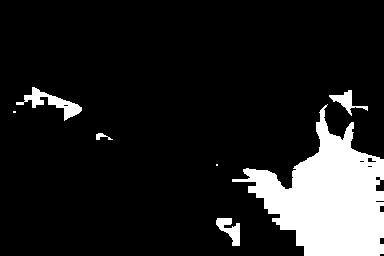}}

\subfigure{\includegraphics[width=\mywidthSS, height=\mywidthSS]{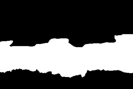}}
\subfigure{\includegraphics[width=\mywidthSS, height=\mywidthSS]{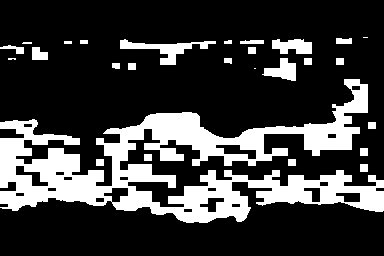}}
\subfigure{\includegraphics[width=\mywidthSS, height=\mywidthSS]{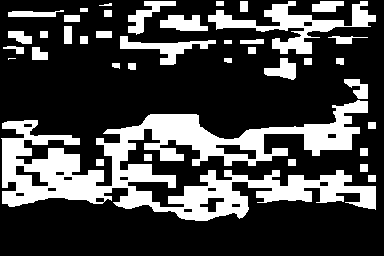}}
\subfigure{\includegraphics[width=\mywidthSS, height=\mywidthSS]{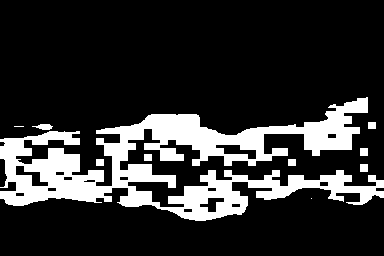}}
\subfigure{\includegraphics[width=\mywidthSS, height=\mywidthSS]{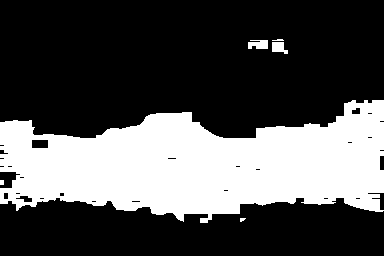}}
\subfigure{\includegraphics[width=\mywidthSS, height=\mywidthSS]{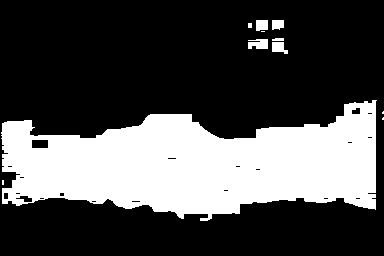}}
\subfigure{\includegraphics[width=\mywidthSS, height=\mywidthSS]{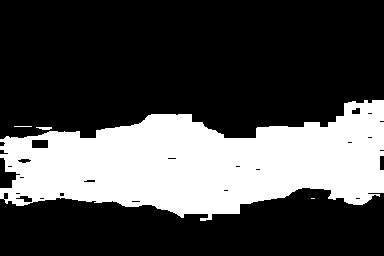}}

\setcounter{subfigure}{0}
\subfigure[Ref label]{\includegraphics[width=\mywidthSS, height=\mywidthSS]{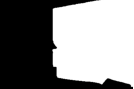}}
\subfigure[$\text{ResNet}_d$]{\includegraphics[width=\mywidthSS, height=\mywidthSS]{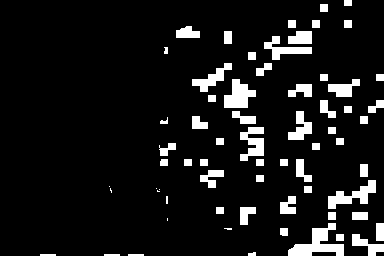}}
\subfigure[$\text{ResNet}_s$]{\includegraphics[width=\mywidthSS, height=\mywidthSS]{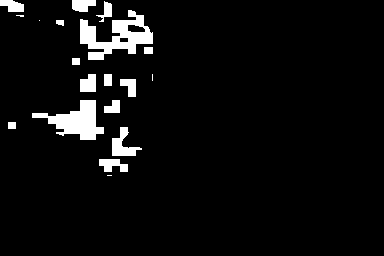}}
\subfigure[$\text{ResNet}_p$]{\includegraphics[width=\mywidthSS, height=\mywidthSS]{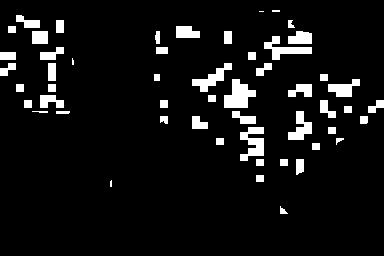}}
\subfigure[$\text{WCBnet}_d$]{\includegraphics[width=\mywidthSS, height=\mywidthSS]{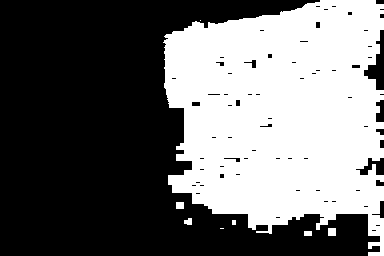}}
\subfigure[$\text{WCBnet}_s$]{\includegraphics[width=\mywidthSS, height=\mywidthSS]{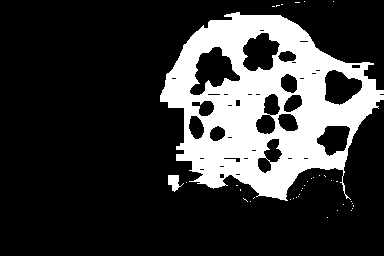}}
\subfigure[$\text{WCBnet}_p$]{\includegraphics[width=\mywidthSS, height=\mywidthSS]{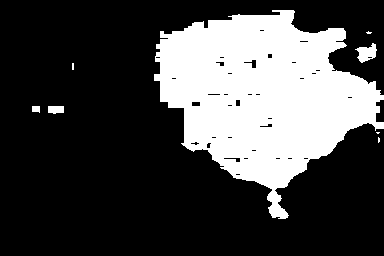}}
\caption{Enhanced heatmaps of several manipulated images,combining activation maps of different extractors and different segmentation maps; The subscript of the model name refers to the associated segmentation model, d for DeepLab, s for SAM, and p for PSPNet.}
\label{fig:enhanced_heatmap}
\end{figure}
We integrate the multi-view activation map \( A \) with the single-class mask \( M_{i^*} \) derived from segmentation maps using Bayesian inference. The resulting enhanced heatmaps \( H \) are presented in Figure~\ref{fig:enhanced_heatmap}. Visually, the segmented regions produced by DeepLab more accurately approximate the size of the manipulations in the testing images of the dataset, making the enhanced heatmap \( \text{WCBnet}_d \) more aligned with the reference label compared to the other heatmaps. In contrast, the finer segmented regions from PSPNet and SAM lead to excessive detail, causing parts of the manipulated region to be excluded by the activation map. Notably, while ResNet50 achieves high image-wise classification accuracy, its pixel-wise localization results misclassify manipulated regions or produce inaccurate maps.

To further assess pixel-wise manipulation localization statistically, we evaluate the enhanced heatmaps using several pixel-wise metrics. The AUC and F1-score are computed to measure the similarity between the generated manipulation localization maps and the ground-truth labels. For this evaluation, we manually selected 40 images that are correctly identified as manipulated and for which the activation maps approximately highlight the target region.
The performance of weakly-supervised WCBnet with different pre-trained segmentation models is also compared to a fully-supervised model name ManTraNet~\cite{wu2019mantra}. 
\begin{table}[t]\scriptsize
\centering
\begin{tabular}{|c|ccc|ccc|c|}
\hline
\multirow{2}{*}{} & \multicolumn{3}{c|}{ResNet}                                        & \multicolumn{3}{c|}{WCBnet}                                        & \multirow{2}{*}{\begin{tabular}[c]{@{}c@{}}ManTraNet~\cite{wu2019mantra}\\ (fully-supervised)\end{tabular}} \\ \cline{2-7}
                  & \multicolumn{1}{c|}{DeepLab} & \multicolumn{1}{c|}{SAM}   & PSPnet & \multicolumn{1}{c|}{DeepLab} & \multicolumn{1}{c|}{SAM}   & PSPnet &                                                                                         \\ \hline \hline
AUC               & \multicolumn{1}{c|}{0.535}   & \multicolumn{1}{c|}{0.545} & 0.524  & \multicolumn{1}{c|}{0.704}   & \multicolumn{1}{c|}{0.608} & 0.703  & 0.653                                                                                   \\ \hline
F1                & \multicolumn{1}{c|}{0.296}   & \multicolumn{1}{c|}{0.257} & 0.259  & \multicolumn{1}{c|}{0.682}   & \multicolumn{1}{c|}{0.310} & 0.365  & 0.238                                                                                   \\ \hline
\end{tabular}
\caption{The pixel-wise performance comparison between the backbone network, proposed method and a fully-supervised ManTraNet; In which the weakly supervised method combines image-wise models and segmentation models.}
\label{tab:PL_heatmap}
\end{table}
Table~\ref{tab:PL_heatmap} compares the pixel-wise performance of two weakly-supervised methods, ResNet and WCBnet, with the fully-supervised ManTraNet. WCBnet outperforms the backbone across all metrics, particularly with the DeepLab model, where it achieves an AUC of 0.704 and an F1-score of 0.682, compared to ResNet's 0.535 and 0.296, respectively. When compared to a fully-supervised method ManTraNet, WCBnet that is enhanced by DeepLab surpasses ManTraNet in F1-score (0.682 over 0.238) and also in AUC (0.704 and 0.653). The results demonstrate WCBnet's superior balance between precision and recall in weakly-supervised scenarios, and prove the feasibility of image manipulation localization without pixel-wise annotations.


\section{Conclusions}
\label{sec:conclusion}
In this paper, we have proposed a novel method for localizing image manipulations without requiring pixel-level labels by combining activation maps generated from image-level manipulation detection networks with the image segmentation maps from pre-trained segmentation models. To achieve this, we integrated class activation maps, incorporating different feature fusion structures named CBAM across various receptive fields and computing geometric mean of these multi-view activation maps to obtain heatmaps. By leveraging Bayesian inference to combine multi-view heatmaps with the segmented region mask from pre-trained segmentation networks, we have achieved an improvement in F1 scores for manipulation localization, enhancing performance by 5\% to 11\% compared to the backbone model. The significance of our results lies in demonstrating the feasibility of localizing image manipulations without relying on pixel-level labels, which is a departure from existing manipulation localization models. Addressing the limitations of small-region manipulation and improving the model’s robustness against various manipulation scenarios will be the focus of our future research.

\bibliography{egbib}

\begin{thebibliography}{26}
\providecommand{\natexlab}[1]{#1}
\providecommand{\url}[1]{\texttt{#1}}
\expandafter\ifx\csname urlstyle\endcsname\relax
  \providecommand{\doi}[1]{doi: #1}\else
  \providecommand{\doi}{doi: \begingroup \urlstyle{rm}\Url}\fi

\bibitem[Abir et~al.(2023)Abir, Khanam, Alam, Hadjouni, Elmannai, Bourouis,
  Dey, and Khan]{abir2023detecting}
Wahidul~Hasan Abir, Faria~Rahman Khanam, Kazi~Nabiul Alam, Myriam Hadjouni,
  Hela Elmannai, Sami Bourouis, Rajesh Dey, and Mohammad~Monirujjaman Khan.
\newblock Detecting deepfake images using deep learning techniques and
  explainable {AI} methods.
\newblock \emph{Intelligent Automation \& Soft Computing}, 35\penalty0
  (2):\penalty0 2151--2169, 2023.

\bibitem[Barni et~al.(2021)Barni, Phan, and Tondi]{9298821}
Mauro Barni, Quoc-Tin Phan, and Benedetta Tondi.
\newblock Copy move source-target disambiguation through multi-branch {CNNs}.
\newblock \emph{IEEE Transactions on Information Forensics and Security},
  16:\penalty0 1825--1840, 2021.
\newblock \doi{10.1109/TIFS.2020.3045903}.

\bibitem[Camacho and Wang(2022)]{camacho2022convolutional}
Ivan~Castillo Camacho and Kai Wang.
\newblock Convolutional neural network initialization approaches for image
  manipulation detection.
\newblock \emph{Digital Signal Processing}, 122:\penalty0 103376, 2022.

\bibitem[Chennamma and Madhushree(2023)]{survey1}
H.~R. Chennamma and B.~Madhushree.
\newblock A comprehensive survey on image authentication for tamper detection
  with localization.
\newblock \emph{Multimedia Tools and Applications}, 82\penalty0 (2):\penalty0
  1873--1904, 01 2023.

\bibitem[Dong et~al.(2013)Dong, Wang, and Tan]{6625374}
Jing Dong, Wei Wang, and Tieniu Tan.
\newblock {CASIA} image tampering detection evaluation database.
\newblock In \emph{2013 IEEE China Summit and International Conference on
  Signal and Information Processing}, pages 422--426, 2013.
\newblock \doi{10.1109/ChinaSIP.2013.6625374}.

\bibitem[Gunawan et~al.(2017)Gunawan, Hanafiah, Kartiwi, Ismail, Za’bah, and
  Nordin]{gunawan2017development}
Teddy~Surya Gunawan, Siti Amalina~Mohammad Hanafiah, Mira Kartiwi, Nanang
  Ismail, Nor~Farahidah Za’bah, and Anis~Nurashikin Nordin.
\newblock Development of photo forensics algorithm by detecting photoshop
  manipulation using error level analysis.
\newblock \emph{Indonesian Journal of Electrical Engineering and Computer
  Science}, 7\penalty0 (1):\penalty0 131--137, 2017.

\bibitem[Guo et~al.(2023)Guo, Liu, Ren, Grosz, Masi, and Liu]{Guo_2023_CVPR}
Xiao Guo, Xiaohong Liu, Zhiyuan Ren, Steven Grosz, Iacopo Masi, and Xiaoming
  Liu.
\newblock Hierarchical fine-grained image forgery detection and localization.
\newblock In \emph{Proceedings of the IEEE/CVF Conference on Computer Vision
  and Pattern Recognition (CVPR)}, pages 3155--3165, June 2023.

\bibitem[Gupta(2023)]{gupta2023image}
Divam Gupta.
\newblock Image segmentation {Keras}: Implementation of {Segnet}, {FCN},
  {Unet}, {PSPnet} and other models in {Keras}.
\newblock \emph{arXiv preprint arXiv:2307.13215}, 2023.

\bibitem[Hac{\i}efendio{\u{g}}lu et~al.(2024)Hac{\i}efendio{\u{g}}lu, Adanur,
  and Demir]{haciefendiouglu2024automatic}
Kemal Hac{\i}efendio{\u{g}}lu, S{\"u}leyman Adanur, and G{\"o}khan Demir.
\newblock Automatic landslide segmentation using a combination of grad-{CAM}
  visualization and {K}-means clustering techniques.
\newblock \emph{Iranian Journal of Science and Technology, Transactions of
  Civil Engineering}, 48\penalty0 (2):\penalty0 943--959, 2024.

\bibitem[Hamid et~al.(2023)Hamid, Elyassami, Gulzar, Balasaraswathi, Habuza,
  and Wani]{hamid2023improvised}
Yasir Hamid, Sanaa Elyassami, Yonis Gulzar, Veeran~Ranganathan Balasaraswathi,
  Tetiana Habuza, and Sharyar Wani.
\newblock An improvised {CNN} model for fake image detection.
\newblock \emph{International Journal of Information Technology}, 15\penalty0
  (1):\penalty0 5--15, 2023.

\bibitem[Hasany et~al.(2023)Hasany, Petitjean, and
  M\'eriaudeau]{Hasany_2023_CVPR}
Syed~Nouman Hasany, Caroline Petitjean, and Fabrice M\'eriaudeau.
\newblock {Seg-XRes-CAM}: Explaining spatially local regions in image
  segmentation.
\newblock In \emph{Proceedings of the IEEE/CVF Conference on Computer Vision
  and Pattern Recognition (CVPR) Workshops}, pages 3733--3738, June 2023.

\bibitem[Kirillov et~al.(2023)Kirillov, Mintun, Ravi, Mao, Rolland, Gustafson,
  Xiao, Whitehead, Berg, Lo, Doll{\'a}r, and Girshick]{kirillov2023segany}
Alexander Kirillov, Eric Mintun, Nikhila Ravi, Hanzi Mao, Chloe Rolland, Laura
  Gustafson, Tete Xiao, Spencer Whitehead, Alexander~C. Berg, Wan-Yen Lo, Piotr
  Doll{\'a}r, and Ross Girshick.
\newblock Segment anything.
\newblock \emph{arXiv:2304.02643}, 2023.

\bibitem[Kweon et~al.(2023)Kweon, Yoon, and Yoon]{kweon2023weakly}
Hyeokjun Kweon, Sung-Hoon Yoon, and Kuk-Jin Yoon.
\newblock Weakly supervised semantic segmentation via adversarial learning of
  classifier and reconstructor.
\newblock In \emph{Proceedings of the IEEE/CVF Conference on Computer Vision
  and Pattern Recognition}, pages 11329--11339, 2023.

\bibitem[Selvaraju et~al.(2020)Selvaraju, Cogswell, Das, Vedantam, Parikh, and
  Batra]{selvaraju2020grad}
Ramprasaath~R Selvaraju, Michael Cogswell, Abhishek Das, Ramakrishna Vedantam,
  Devi Parikh, and Dhruv Batra.
\newblock {Grad-CAM}: visual explanations from deep networks via gradient-based
  localization.
\newblock \emph{International journal of computer vision}, 128:\penalty0
  336--359, 2020.

\bibitem[Thakur and Rohilla(2020)]{thakur2020recent}
Rahul Thakur and Rajesh Rohilla.
\newblock Recent advances in digital image manipulation detection techniques: A
  brief review.
\newblock \emph{Forensic science international}, 312:\penalty0 110311, 2020.

\bibitem[Tyagi and Yadav(2023)]{tyagi2023detailed}
Shobhit Tyagi and Divakar Yadav.
\newblock A detailed analysis of image and video forgery detection techniques.
\newblock \emph{The Visual Computer}, 39\penalty0 (3):\penalty0 813--833, 2023.

\bibitem[Walia et~al.(2022)Walia, Kumar, Agarwal, and Kim]{walia2022using}
Savita Walia, Krishan Kumar, Saurabh Agarwal, and Hyunsung Kim.
\newblock Using {XAI} for deep learning-based image manipulation detection with
  shapley additive explanation.
\newblock \emph{Symmetry}, 14\penalty0 (8):\penalty0 1611, 2022.

\bibitem[Wang and Abhayaratne(2023)]{wang2023wcbnet}
Ziyong Wang and Charith Abhayaratne.
\newblock {WCBnet}: Weighted convolutional block modelling of signed-value
  error levels for image-wise copy-move and splicing detection.
\newblock In \emph{2023 IEEE 25th International Workshop on Multimedia Signal
  Processing (MMSP)}, pages 1--6. IEEE, 2023.

\bibitem[Weber et~al.(2021)Weber, Wang, Qiao, Xie, Collins, Zhu, Yuan, Kim, Yu,
  Cremers, Leal-Taixe, Yuille, Schroff, Adam, and Chen]{deeplab2_2021}
Mark Weber, Huiyu Wang, Siyuan Qiao, Jun Xie, Maxwell~D. Collins, Yukun Zhu,
  Liangzhe Yuan, Dahun Kim, Qihang Yu, Daniel Cremers, Laura Leal-Taixe,
  Alan~L. Yuille, Florian Schroff, Hartwig Adam, and Liang-Chieh Chen.
\newblock {{DeepLab2}: A TensorFlow Library for Deep Labeling}.
\newblock \emph{arXiv: 2106.09748}, 2021.

\bibitem[Wu et~al.(2019)Wu, AbdAlmageed, and Natarajan]{wu2019mantra}
Yue Wu, Wael AbdAlmageed, and Premkumar Natarajan.
\newblock {ManTra-Net}: Manipulation tracing network for detection and
  localization of image forgeries with anomalous features.
\newblock In \emph{Proceedings of the IEEE/CVF conference on computer vision
  and pattern recognition}, pages 9543--9552, 2019.

\bibitem[Zanardelli et~al.(2023)Zanardelli, Guerrini, Leonardi, and
  Adami]{zanardelli2023image}
Marcello Zanardelli, Fabrizio Guerrini, Riccardo Leonardi, and Nicola Adami.
\newblock Image forgery detection: a survey of recent deep-learning approaches.
\newblock \emph{Multimedia Tools and Applications}, 82\penalty0 (12):\penalty0
  17521--17566, 2023.

\bibitem[Zeng et~al.(2024)Zeng, Wang, Zhou, Zhang, and Meng]{zeng2024semi}
Qiang Zeng, Hongxia Wang, Yang Zhou, Rui Zhang, and Sijiang Meng.
\newblock Semi-supervised image manipulation localization with residual
  enhancement.
\newblock \emph{Expert Systems with Applications}, 252:\penalty0 124171, 2024.

\bibitem[Zhai et~al.(2023)Zhai, Luan, Doermann, and Yuan]{zhai2023towards}
Yuanhao Zhai, Tianyu Luan, David Doermann, and Junsong Yuan.
\newblock Towards generic image manipulation detection with weakly-supervised
  self-consistency learning.
\newblock In \emph{Proceedings of the IEEE/CVF International Conference on
  Computer Vision}, pages 22390--22400, 2023.

\bibitem[Zhang et~al.(2020)Zhang, Zhou, Zhao, Man, Liu, and
  Yao]{zhang2020survey}
Man Zhang, Yong Zhou, Jiaqi Zhao, Yiyun Man, Bing Liu, and Rui Yao.
\newblock A survey of semi-and weakly supervised semantic segmentation of
  images.
\newblock \emph{Artificial Intelligence Review}, 53:\penalty0 4259--4288, 2020.

\bibitem[Zhou et~al.(2024)Zhou, Wang, Zeng, Zhang, and Meng]{zhou2024exploring}
Yang Zhou, Hongxia Wang, Qiang Zeng, Rui Zhang, and Sijiang Meng.
\newblock Exploring weakly-supervised image manipulation localization with
  tampering edge-based class activation map.
\newblock \emph{Expert Systems with Applications}, 249:\penalty0 123501, 2024.

\bibitem[Ț{\^a}nțaru et~al.(2024)Ț{\^a}nțaru, Oneaț{\u{a}}, and
  Oneaț{\u{a}}]{tantaru2024weakly}
Dragoș-Constantin Ț{\^a}nțaru, Elisabeta Oneaț{\u{a}}, and Dan
  Oneaț{\u{a}}.
\newblock Weakly-supervised deepfake localization in diffusion-generated
  images.
\newblock In \emph{Proceedings of the IEEE/CVF Winter Conference on
  Applications of Computer Vision}, pages 6258--6268, 2024.

\end{thebibliography}
\end{document}